\documentclass[10pt,twocolumn,letterpaper]{article}

\usepackage[pagenumbers]{cvpr} 

\newcommand{\ours}[0]{\textsc{PRISM}}
\newcommand{\cole}[0]{\texttt{OpenCole}}
\newcommand{\promptDiv}[0]{\texttt{Prompt2Diverse}}
\newcommand{\dataOne}[0]{\texttt{Data2One}}
\newcommand{\dataDiv}[0]{\texttt{Data2Diverse}}

\usepackage{amsmath}
\usepackage{amssymb}
\usepackage{mathtools}
\usepackage{amsthm}
\usepackage{cuted}   
\usepackage{capt-of} 
\usepackage{lipsum} 
\usepackage{booktabs}
\usepackage[table]{xcolor}

\usepackage[toc,page,header]{appendix}
\usepackage{minitoc}

\usepackage{algorithm}
\usepackage{algorithmic}
\usepackage{transparent}
\definecolor{perfblue}{RGB}{64, 114, 175}
\newcommand{\algcommentlight}[1]{\textcolor{perfblue}{\transparent{0.8}\small{\texttt{\textbf{//\hspace{2pt}#1}}}}}

\usepackage{listings}

\definecolor{cvprblue}{rgb}{0.21,0.49,0.74}
\usepackage[pagebackref,breaklinks,colorlinks,allcolors=cvprblue]{hyperref}

\def\paperID{*****} 
\def\confName{CVPR}
\def\confYear{2026}

\title{PRISM: Learning Design Knowledge from Data \\ for Stylistic Design Improvement}

\author{
Huaxiaoyue Wang$^1$, Sunav Choudhary$^2$, Franck Dernoncourt$^2$, Yu Shen$^2$, Stefano Petrangeli$^2$ \\
$^1$Cornell University, $^2$Adobe Research \\
yukiwang@cs.cornell.edu, \{schoudha, dernonco, shenyu, petrange\}@adobe.com
}

\begin{document}
\doparttoc 
\faketableofcontents 

\maketitle
\begin{abstract}
Graphic design often involves exploring different stylistic directions, which can be time-consuming for non-experts.
We address this problem of stylistically improving designs based on natural language instructions. 
While VLMs have shown initial success in graphic design, their pretrained knowledge on styles is often too general and misaligned with specific domain data.
For example, VLMs may associate minimalism with abstract designs, whereas designers emphasize shape and color choices. 
Our key insight is to leverage design data---a collection of real-world designs that implicitly captures designer's principles---to learn design knowledge and guide stylistic improvement. 
We propose \ours{} (PRior-Informed Stylistic Modification) that constructs and applies a design knowledge base through three stages: (1) clustering high-variance designs to capture diversity within a style, (2) summarizing each cluster into actionable design knowledge, and (3) retrieving relevant knowledge during inference to enable style-aware improvement. 
Experiments on the Crello dataset shows that \ours{} achieves the highest average rank of $1.49$ (closer to 1 is better) over baselines in style alignment. User studies further validate these results, showing that \ours{} is consistently preferred by designers. 
\end{abstract}
    
\section{Introduction}
\label{sec:intro}
\begin{figure*}[ht]
    \centering
    \includegraphics[width=\textwidth]{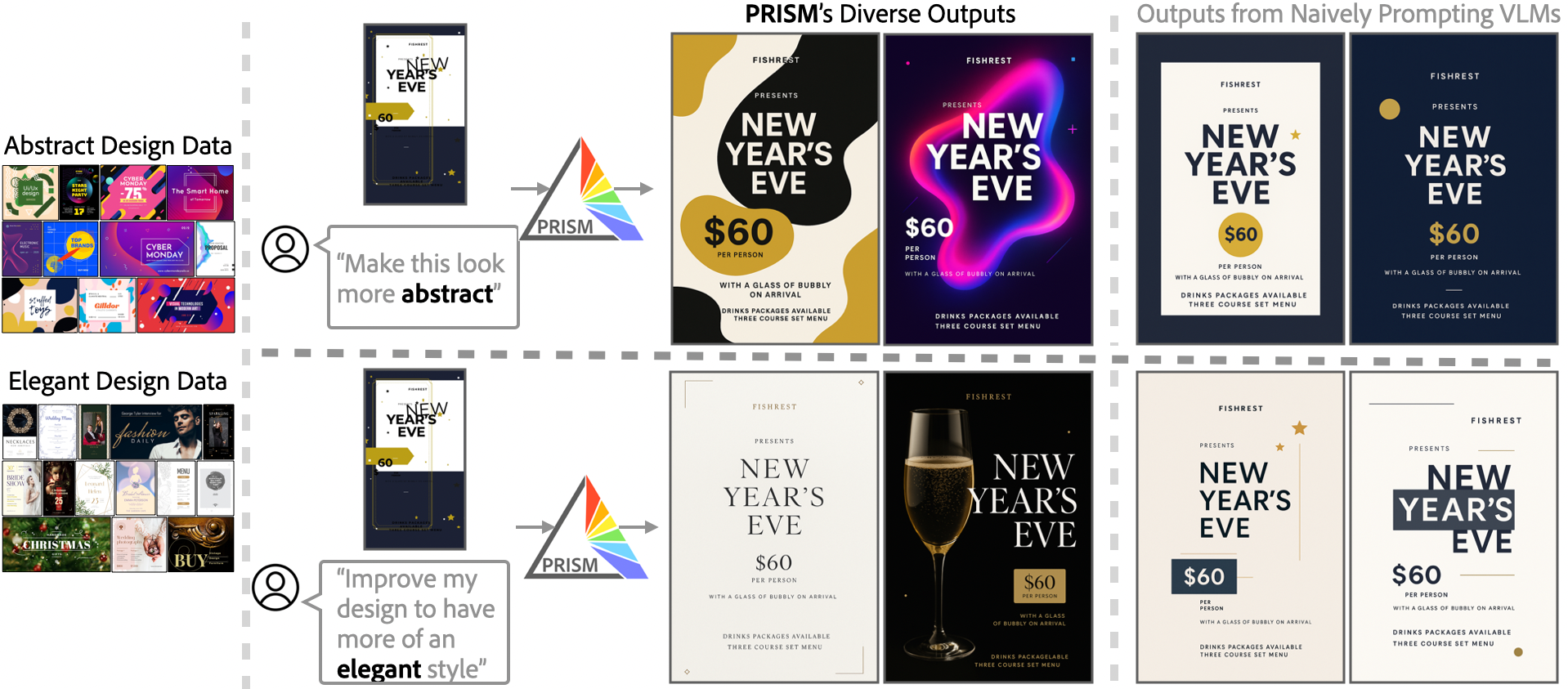}
    \caption{A user provides a design to improve and a natural language instruction. \ours{} leverages existing design data to generate diverse improvements that align with the requested style. In contrast, approaches that solely rely on VLMs' pretrained knowledge on styles produce outputs that fail to match the design data.}
    \label{fig:problem_formulation}
\end{figure*}
Graphic design serves as a powerful form of visual communications, where stylistic choices influence mood and perception. 
Recent advances have enabled everyday users to generate visually appealing designs from natural language instructions and assets~\cite{jia2024colehierarchicalgenerationframework, inoue2024opencolereproducibleautomaticgraphic, wang2025banneragencyadvertisingbannerdesign}. 
Yet, in practice, graphic design is rarely a one-shot process. Designers often iterate through multiple stylistic directions, a task that remains time-consuming and challenging for non-experts.
Thus, the ability to improve designs stylistically based on language instructions, as shown in Fig.~\ref{fig:problem_formulation}, is equally important. 

Computational works for graphic designs have studied layout generation~\cite{yamaguchi2021canvasvae, li2021attribute-conditioned, talton2012designpatterns, hsu2023posterlayout, gupta2021layouttransformer, Horita_2024_CVPR, ijcai2022p692}, design completion~\cite{kikuchi2025md, gupta2021layouttransformer}, typography synthesis~\cite{gao2023textpainter, Shimoda_2024_WACV}, and more. 
A closely related area is design improvement---judging designs and improving layouts for aesthetics.
Existing methods primarily focus on spatially rearranging elements to improve visual appeal, but they do not condition such edits on design styles~\cite{aesthetics2023, murray2012ava, haraguchi2024gpteval, lu2014rapid, designometer2025, nag2025agenticdesignreview}. 
Meanwhile, recent works demonstrate the effectiveness of leveraging pretrained knowledge in Vision Language Models (VLMs) for layout generation~\cite{cheng2025graphic, seol2024posterllama, lin2023layoutprompter, yang2024posterllavaconstructingunifiedmultimodal, patnaik2025aesthetiq}, opening opportunities for language-conditioned design tasks. 
However, Fig.~\ref{fig:problem_formulation} shows that directly prompting VLMs for stylistic improvements often produces edits that lack details and diversity, even though they may implicitly encode design principles during pretraining. 
We posit that this limitation stems from the VLMs' pretrained knowledge being too general and lacking alignment with specific domain data. 

Our key insight is to \textbf{\emph{leverage existing design data to explicitly learn design knowledge}} and use it to guide stylistic improvement. 
Collections of real-world designs often include designer-provided tags such as ``abstract" or ``elegant".
Each design reflects the designers' underlying stylistic principles, providing a rich source for learning design knowledge.
However, directly using these designs as input to VLMs is ineffective because designs sharing the same tag can still significantly differ from each other visually as shown in Fig~\ref{fig:problem_formulation}, making it challenging to extract fine-grained, actionable knowledge. 

To address this, we propose \ours{} (PRior-Informed Stylistic Modification), a framework that effectively learns and utilizes a design knowledge base via a three-stage process: style space partitioning, style knowledge extraction, and prior-informed edits at inference time. 
\emph{(1) Partitioning the Style Space:} In order to handle variances and capture diversity within a style, we compute pairwise distances that capture semantic and spatial relationships between all designs before identifying visually coherent clusters via K-medoids. 
\emph{(2) Extracting Style Knowledge:} With all the clusters, the goal is to learn a set of compact, actionable knowledge that can be used by the downstream design improvement module. Thus, we employ a contrastive summarization process that aligns extracted knowledge with designs in the same cluster while remaining discriminative from others. 
\emph{(3) Prior-Informed Editing During Inference: } To make improvements that align with the original data during inference, we integrate learned knowledge as priors into a Retrieval-Augmented Generation (RAG) pipeline, enabling proportional retrieval of relevant design knowledge for style-aware edits. 
Our key contributions are:
\begin{enumerate}[leftmargin=*]
    \setlength{\itemsep}{0pt}
    \item A method for leveraging design data to construct a design knowledge base for design improvement. We aggregate the data into meaningful subsets before distilling each into concise, distinctive design knowledge.
    \item An inference mechanism for style-aware improvement that retrieves appropriate design knowledge proportionally to the original data distribution. 
    \item Experiments on the Crello dataset showing that \ours{} achieves a highest average style alignment ($0.847 \rightarrow 0.999)$ compared to the best baseline. User studies validate this result, and show that designers prefer \ours{} across multiple design axis.  
\end{enumerate}

\section{Related Works}
\label{sec:related_works}
\textbf{Graphic Design Improvement.} 
Fundamentally, we tackle the problem of design improvement, which inherently involves evaluating the current design. 
Prior works have solely focused on design evaluation using heuristics~\cite{DavidChekLingNgo2000, Ngo2001, harrington2004aesthetic, BAUERLY2006670, zen2014towards, o2014learning}, deep learning~\cite{murray2012ava, lu2014rapid, zhao2018personalities, tabata2019autolayout} or large pretrained models such as VLMs~\cite{haraguchi2024gpteval, nag2025agenticdesignreview}. 
Other works aim to enhance a design's aesthetics by improving layouts~\cite{pang2016visualflowweb, rahman2021ruite, aesthetics2023, zhang2024vascar, designometer2025, iwai2024layout}.
However, their methods primarily adjust the assets' positions and sizes and do not condition edits on specific design style. 
In contrast, our work focuses on stylistic design improvement by leveraging learned design knowledge to edit colors, fonts, assets, and layouts in order to achieve the requested style.

\begin{figure*}[ht]
    \centering
    \includegraphics[width=\textwidth]{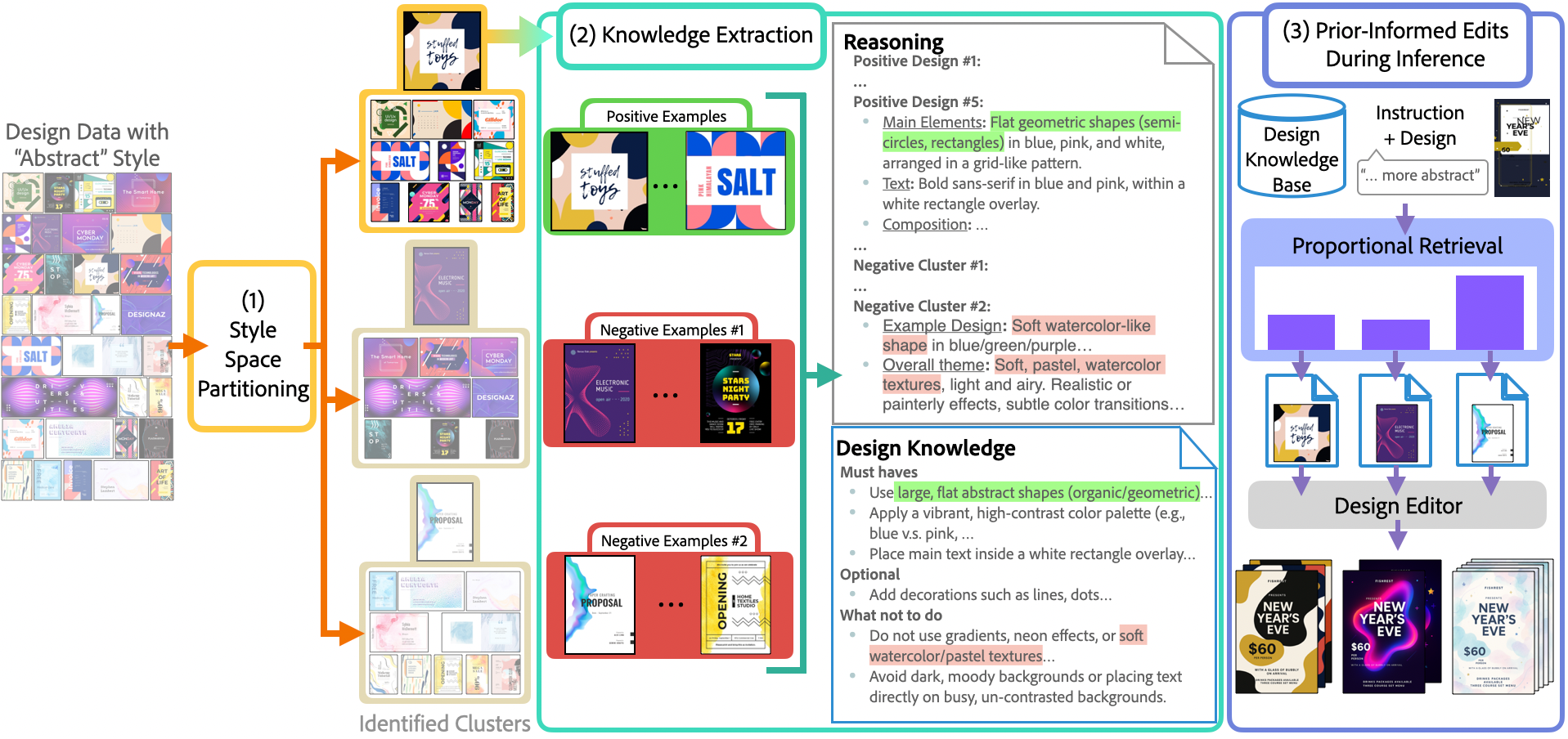}
    \caption{\textbf{\ours{} Overview.} Given design data with a style ``abstract," the (1) Style Space Partitioning stage identifies visually distinctive clusters. Then, (2) Knowledge Extraction focuses on learning concise, actionable knowledge from these clusters.
    Under a contrastive framework, the VLM compares and contrast positive and negative examples before generating the knowledge.
    We highlight two examples of how positive/negative designs influence specific lines in the guideline.
    During inference, (3) Prior-Informed Edits proportionally retrieves relevant design knowledge based on the original design data distribution, thereby outputting diverse improvements that also align with data.}
    \label{fig:stage1and2}
\end{figure*}

\noindent \textbf{LLMs and VLMs for Graphic Design.} 
LLMs and VLMs have shown initial success in generating layouts given assets~\cite{lin2023layoutprompter, chen2023reasonlayoutevokinglayout, seol2024posterllama, yang2024posterllavaconstructingunifiedmultimodal, cheng2024colaycontrollablelayoutgeneration, tang2024layoutnuwa, patnaik2025aesthetiq, cheng2025graphic, wang2025sega, anonymous2025anylayout}, but their generation process is not steered by design styles. 
The closest works to ours are VLM-based systems that construct complete designs given a combination of natural language and assets~\cite{jia2024colehierarchicalgenerationframework, inoue2024opencolereproducibleautomaticgraphic, Qu2025IGD, zhang2025creatipostereditablecontrollablemultilayer, wang2025banneragencyadvertisingbannerdesign}. These systems typically adopt a hierarchical pipeline, where an image diffusion model generates the background and other modules handle text and layout.
Despite supporting mult-modal input, their ability to produce diverse design styles remains limited.  
For example, IGD~\cite{Qu2025IGD} can only condition its background on imitating a single reference image, while CreatiPoster~\cite{zhang2025creatipostereditablecontrollablemultilayer} and BannerAgency~\cite{wang2025banneragencyadvertisingbannerdesign} heavily rely on detailed textual descriptions of stylistic elements, increasing the prompting burden for everyday users.
Our work complements these approaches by enriching the VLM-based systems with learned design knowledge, enabling stylistic improvement that aligns with the provided design data. 

\noindent\textbf{Knowledge-Augmented Generation.}
Retrieval-augmented generation improves LLM performance by augmenting the input with relevant information~\cite{lewis2020rag, guu2020rag}. 
Our work is closely related to approaches that automatically construct knowledge bases for retrieval~\cite{qiao2024agent, fu2024autoguide, zhao2024expel, nekoei2025justintimeepisodicfeedbackhinter, tang2025agent}.
Some works in LLM agents generate knowledge given a pair of positive and negative trajectories.
To obtain these pairs, ~\citeauthor{qiao2024agent} requires a simulator to collect agent trajectories as negative examples, while others~\cite{fu2024autoguide,zhao2024expel, nekoei2025justintimeepisodicfeedbackhinter,tang2025agent} also depend on reward functions to identify high-reward trajectories as positive examples. 
Another line of work explicitly train high-level planners that generate guidelines to steer a frozen low-level policy~\cite{xiong-etal-2025-mpo, si2025goalplanjustwish}, but they also require execution reward signals for optimization. 
These strategies are impractical for design tasks, as we lack simulators and ground-truth reward functions that reflect real-world distributions of user-provided designs. 
In contrast, our work only leverages an easily acquired collection of high-quality designs and automatically curates positive and negative sets to learn meaningful design knowledge.
\section{Approach}
We introduce \ours{} (Prior-Informed Stylistic Modification) to address the test-time adaptation problem of aligning VLMs to domain-specific design data and enabling style-aware design improvements. 
We assume access to a collection of designs, each annotated with stylistic tags (e.g., ``abstract", ``elegant").
This type of data is accessible from benchmarks (e.g,. Crello~\cite{yamaguchi2021canvasvae}) or available as proprietary datasets. 
\ours{} consists of three main stages to leverage this design data, shown in Fig~\ref{fig:stage1and2}. In order to handle variances within a style, stage 1 explicitly identify meaningful clusters. Then, each cluster from stage 1 needs to be captured in a concise, transferrable form, so stage 2 explicitly learns a design knowledge via a contrastive learning framework. To effectively utilize the learned knowledge during inference, stage 3 proportionally retrieves the appropriate design data, enabling design improvements that matches the distribution of the provided design data. 

\subsection{Style Space Partitioning\label{sec:approach_data_curation}}
Designs that share the same stylistic tag still have highly varied visuals. 
Thus, this stage aims to handle this variance and capture diversity within a style by organizing the data into visually coherent clusters.  
This process requires an effective visual distance metric to measure similarity between two designs and a suitable clustering algorithm. 

To capture both semantic and spatial relationships between designs, we adopt the GRAph-based Design (GRAD) distance proposed by~\citeauthor{nag2025agenticdesignreview}.
The GRAD distance represents each design as a graph, where each vertex corresponds to an image patch and each edge is the cosine distance between patch embeddings from image encoders such as CLIP~\cite{radford2021learning} or DINOv2~\cite{oquab2023dinov2}.
By applying optimal transport to match these graphs, the GRAD distance captures fine-grained structural similarities that distance approaches based on global image embeddings often miss. 

For clustering, we first aggregate all the designs that have the same style by their stylistic tags before computing the pairwise GRAD distances between all of them.  
Then, we cluster the designs via the K-medoids algorithm~\cite{rdusseeun1987clustering}, where $K$ controls the number of clusters.
Each cluster is represented by its medoid---the design closest to all the others in the cluster.
We sweep $K$ (empirically from 2 to 5) and select the value that yields the highest silhouette score, which measures the cluster quality~\cite{rousseuw1987silhouettes}.

\subsection{Style Knowledge Extraction\label{sec:approach_knowledge_extraction}}
Each cluster identified during the style space partitioning stage is distilled into a concise design knowledge for downstream tasks. 
This knowledge must satisfy two desiderata. 
First, it should be actionable, offering concrete example attributes for elements such as text and shapes. 
Second, it should be discriminative, capturing a unique type of visual characteristics within the style. 

We introduce a contrastive learning framework. 
Given a cluster to learn, we sort the designs by their distances to the medoid and select $i$ positive examples (including the medoid and its nearest neighbors). 
Other clusters under the same stylistic tag provide discriminative signals, so we construct the negative examples under the same scheme, using $j$ negative examples. 
Empirically, we set $i=25$ and $j=10$. 


Conditioning on these visual examples, the VLM first analyzes the positive designs in detail to identify recurring themes before examining for broader patterns.
It repeats the process for the negative examples, focusing on features that discriminate the positives from the negatives. 
Fig.~\ref{fig:stage1and2} shows example reasoning outputs during this stage.
Finally, the VLM synthesizes its finding into the design knowledge, containing must-have features that capture what most positive designs have in common (e.g,. ``use large, flat abstract shapes (organic/geometric)"), optional attributes that describes what some positive designs share (e.g., ``add decorations such as lines, dots..."), and must-not-have decisions that highlight what makes negative designs different from the positive ones (e.g., ``do not use gradients, neon effects, or soft watercolor..."). 

After generating the knowledge, we summarize each document into a single-sentence via an LLM for efficient retrieval. 
The resulting knowledge base indexes each entry by its style tag, summary, and the detailed guidelines. 

\subsection{Prior-Informed Edits}
At inference time, the user provides a suboptimal design and a language instruction. 
Building on Retrieval-Augmented Generation (RAG), \ours{} first identifies the design styles that align with the instruction. 
Depending on how many variations the user wants, \ours{} retrieve the design knowledge differently. 
If only one is requested, we compute the Sentence-BERT~\cite{reimers2019sentencebertsentenceembeddingsusing} embedding of the instruction combined with a caption of the provided design, then retrieve the knowledge with the closest summary.
If multiple variations are needed, we sample knowledge proportionally to the cluster size, ensuring outputs reflect the original data distribution. 
For the downstream design improvement, we adapt the hierarchical pipeline in prior works COLE~\cite{jia2024colehierarchicalgenerationframework} and OpenCole~\cite{inoue2024opencolereproducibleautomaticgraphic}.
The retrieved design knowledge is incorporated into the prompt of a VLM planner that generates a concrete design plan based on the original design and user instructions. 
This plan is then used by an image diffusion model to produce the edited design. 

\section{Quantitative Experiments}
\begin{figure*}[ht]
    \centering
    \includegraphics[width=\textwidth]{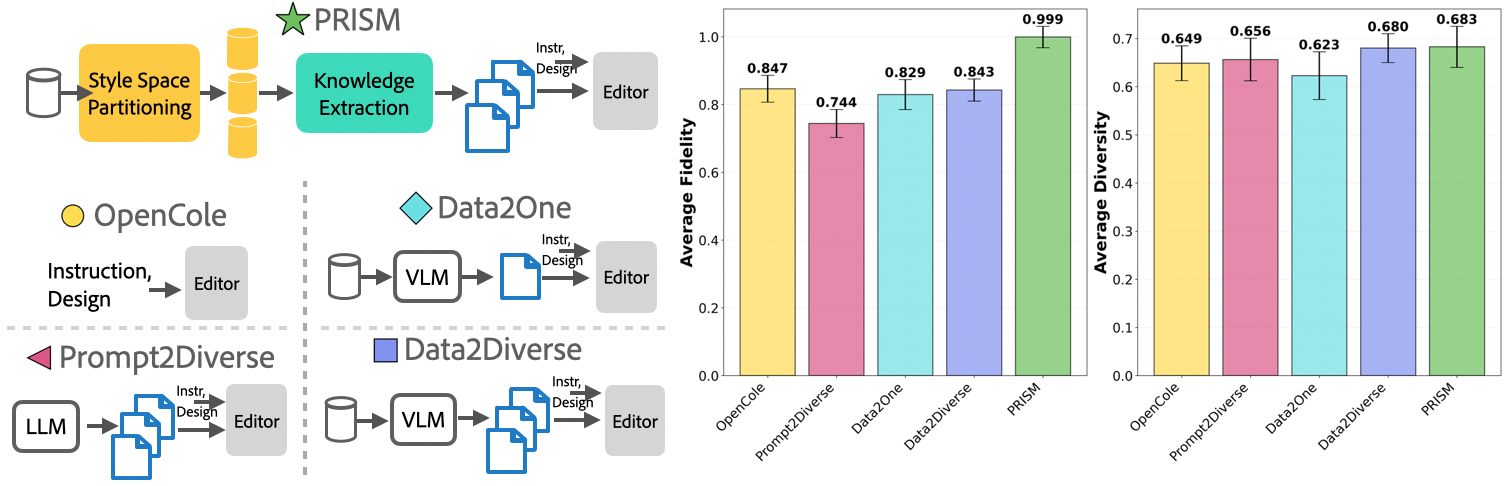}
    \caption{\textbf{Main Results.} We report the average fidelity and diversity with standard error across 15 styles. \ours{} achieve the highest value for both metrics. On the left, we also visualize different methods' input/output.}
    \label{fig:main_results}
\end{figure*}
\begin{figure*}[ht]
    \centering
    \includegraphics[width=\textwidth]{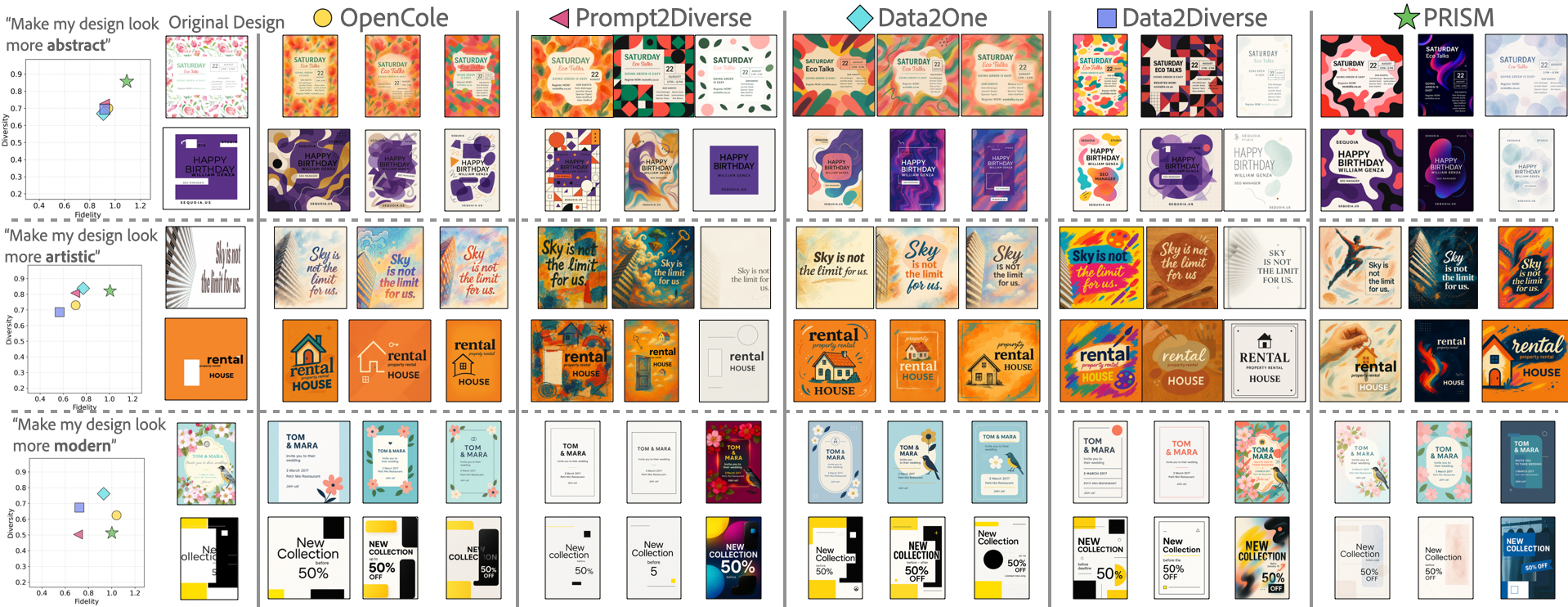}
    \caption{\textbf{Qualitative Results.} We show 3 styles, each representative of \ours{}'s different performance. ``abstract" is one where \ours{} achieves the highest rank for both fidelity and diversity. ``artistic" is where one is high (fidelity). ``modern" is where both ranks are lower. }
    \label{fig:qual}
\end{figure*}
\subsection{Setup}
\textbf{Datasets.}
We evaluate \ours{} on the benchmark graphic design dataset Crello~\cite{yamaguchi2021canvasvae}\footnote{Original images from Crello dataset by Yamaguchi et al. available at \url{https://huggingface.co/datasets/cyberagent/crello}, licensed under CDLA-Permissive-2.0 license}, which contains 19479 designs for training, 1852 for validation, and 1971 for test. 
We combine the training and validation set to be the reference design data.
Each design has a list of associated keywords, which we use to identify the design style.
There are 15 major design styles, each containing at least 100 designs, that our approach should align to: abstract, artistic, bright, cheerful, classic, colorful, corporate, dynamic, elegant, floral, geometric, graphic, modern, natural, simple. 
With these styles, we evaluate how well a method is able to improve a design given a natural language instruction, such as ``make my design look more abstract."
We expect that a performant method would output stylistic designs that resemble the data. 
Details on how we identify the design style from keywords and construct the design data are in Appendix.

\begin{figure}
    \centering
    \includegraphics[width=0.8\linewidth]{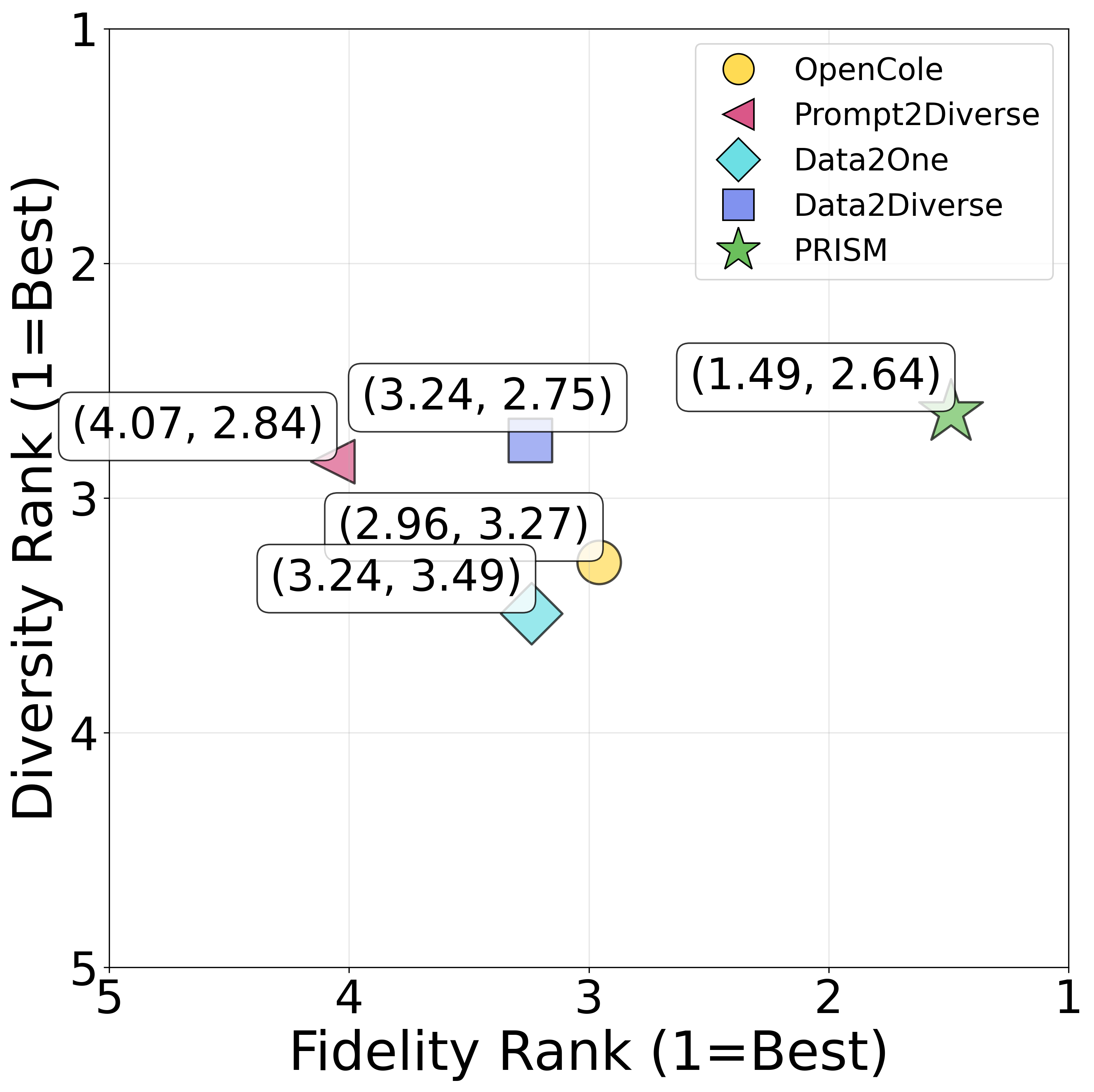}
    \caption{\textbf{Expected Ranks.} We report the expected ranks across all styles. \ours{} achieves the best rank in both fidelity and diversity, showing its ability to balance the two metrics. }
    \label{fig:rank}
\end{figure}

\noindent\textbf{Baselines.} 
Fig.~\ref{fig:main_results} visualizes the input and output of the baselines. 
We compare \ours{} against prior work \textbf{\cole{}}~\cite{inoue2024opencolereproducibleautomaticgraphic}, which we slightly modify to be able to improve designs instead of only generating new designs. 
In addition to language instructions, it now also conditions on the user's original design to generate a design plan before using that plan to edit the design via an image diffusion model. 
Note that \cole{} does not explicitly receive any design knowledge as input.
To test our hypothesis that LLMs' internal knowledge is misaligned from design data, we compare against \textbf{\promptDiv{}}, which directly prompts an LLM to generate $K$ diverse design knowledge about a design style without accessing any data. 
Here, $K$ matches the number of design knowledge that \ours{} generates for the same style. 
For baselines that use design data, we compare against \textbf{\dataOne{}} and \textbf{\dataDiv{}}.
When learning knowledge for a style, they only randomly sample $i=25$ examples instead of identifying meaningful design clusters like \ours{}.
This set up also means that both approaches lack \ours{}'s contrastive learning framework because they only have positive examples. 
In terms of outputs, \dataOne{} learns one design knowledge per style, while \dataDiv{} tries to infer $K$ diverse design knowledge, where $K$ matches \ours{}'s for a style.

\noindent\textbf{Implementation Details.}
For all approaches, we use \texttt{gpt-4.1} to extract design knowledge and generate design plans. 
We use \texttt{gpt-image-1} for making the design improvement conditioning on the design plan and the original design. 
For \ours{}'s data curation step, the GRAD distances is computed using CLIP~\cite{radford2021learning} embeddings from \texttt{CLIP-ViT-L-14-laion2B-s32B-b82K} model provided by OpenCLIP~\cite{ilharco_gabriel_2021_5143773}.
For each design style, the test set is constructed by sampling $3$ test design and $3$ programmatically perturbed test design from 3 other design styles. This results in 18 test designs to improve per style. For each test design, all approaches generate the same amount of design improvements, resulting in around 180 improved designs per style.
More details are in Appendix. 

\noindent \textbf{Metrics.} 
For each style in the design data, there is a collection of $N$ real designs $\{X_i\}$.  
Each method also has a corresponding set of $M$ generated design improvements $\{Y_j\}$. 
Note that $M$ is the same size across all approaches for a style for fair comparisons. 
We use the fidelity and diversity metric proposed by~\citeauthor{naeem2020reliablefidelity} to evaluate how well the two real and generated distribution matches. 
This metric relies on a visual distance metric $d(\cdot, \cdot)$ to measure the distance between two designs, so we set it to be the same GRAD distance used in Section~\ref{sec:approach_data_curation}. 
\textbf{Fidelity} measures ``the degree to which the generated samples resemble the real ones" and concretely ``counts how many real-sample neighborhood spheres contain $Y_j$", so
\begin{equation}
    \text{fidelity} := \frac{1}{kM} \sum_{j=1}^{M} \sum_{i=1}^{N} \mathbf{1}[d(Y_j, X_i) \leq d(X_i, \text{NN}_k(X_i))]
\end{equation}
where $\text{NN}_k(X_i)$ is the $k$-th closet real design to $X_i$. 
\textbf{Diversity} measures ``whether the generated samples cover the full variability of the real samples" and concretely calculates ``the
ratio of real samples that are covered by the fake samples", so
\begin{equation}
    \text{diversity} := \frac{1}{N} \sum_{i=1}^{N} \mathbf{1}[\exists j \text{ s.t. } d(Y_j, X_i) \leq d(X_i, \text{NN}_k(X_i))].
\end{equation}
Because each design style has different number of designs $N$, we unify $k$, which controls the nearest neighbor, to be $k=\alpha N$ and $\alpha=0.05$. 
For all reported results, we perform 10,000 bootstrap resamples of the real and generated data to obtain a robust estimate of the expected performance.

\subsection{Overall Results}
Fig.~\ref{fig:main_results} reports the average fidelity and diversity, and Fig.~\ref{fig:qual} shows some qualitative results.
We also compute the expected rank to access the relative performance in Fig.~\ref{fig:rank}.
Overall, \ours{} outperforms other methods in balancing the trade-off between fidelity and diversity. 

\noindent \textbf{Fidelity.} \ours{} achieves the highest fidelity of 0.999 and the best average rank of 1.49/5, where 1 represents always being the best, showing that \ours{} consistently outperforms baselines in guiding design improvements that resemble the provided data.
Meanwhile, \promptDiv{} performs the worse, with an average fidelity of 0.744 and rank of 4.07/5. Its poor performance suggests that directly prompting the LLM to generate variants of design knowledge within a style leads to greater misalignment from real-world design data. 
Interestingly, although the \cole{} baseline does not explicitly inject any design knowledge during downstream design improvement, its performance is comparable to \dataOne{} and \dataDiv{}, which both have access to data. We closely analyze this phenomenon in Section~\ref{sec:exp_fidelity_discussion}.

\noindent \textbf{Diversity.} \ours{} achieves the highest diversity of 0.683 and the best rank of 2.64/5 with \dataDiv{} being a close second. 
Comparing the rankings reveals two distinct performance groups.
Methods that explicitly inject different variants of design knowledge---\ours{}, \dataDiv{}, \promptDiv{}---achieve diversity ranks of 2.64, 2.75, 2.84 respectively.
In contrast, methods that either omit design knowledge or inject only one design knowledge per style (\cole{}, \dataOne{}) obtain diversity ranks of 3.27 and 3.49 respectively.
The existence of these two groups support our hypothesis that explicitly steering downstream tasks with design knowledge leads to more diverse outputs. 
In addition, leveraging the design data helps lead to enhances diversity by providing signals aligned with real-world distributions.  

\begin{figure*}[ht]
    \centering
    \includegraphics[width=0.9\textwidth]{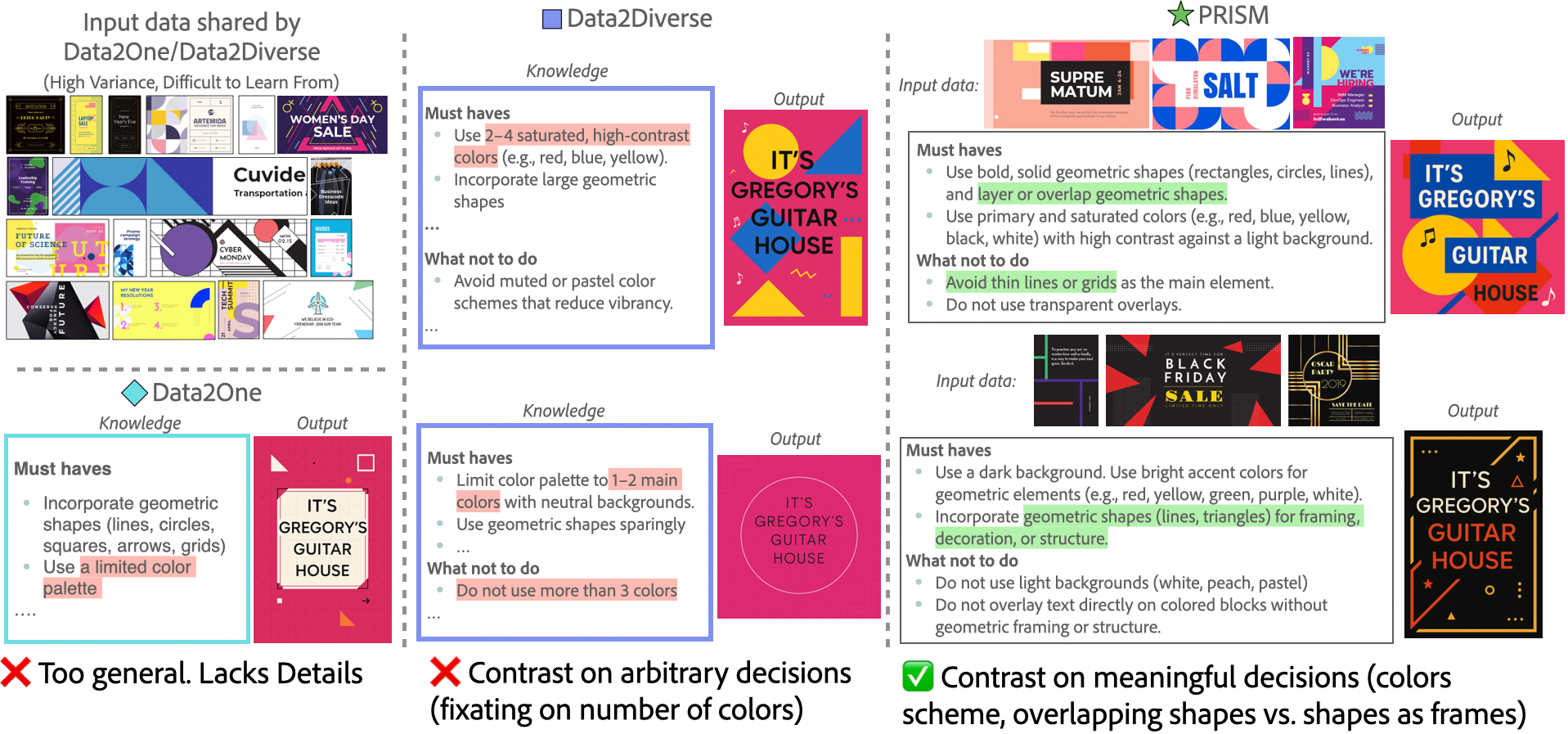}
    \caption{\textbf{Qualitative Comparison of Knowledge Learned by Different Data-Driven Approaches.} Baselines \dataOne{} and \dataDiv{} use randomly sample designs as input, which results in high variance. Conditioning on these, \dataOne{} learns a design knowledge that is too general and lacks detail. \dataDiv{} infers subsets from the noisy data, resulting in knowledge that contrast on arbitrary decisions. In contrast, \ours{} receive inputs that have less variance because of its data curation step, allowing it to learn meaningfully diverse knowledge that results in diverse outputs. }
    \label{fig:fidelity_example}
\end{figure*}
\begin{figure}
    \centering
    \includegraphics[width=\linewidth]{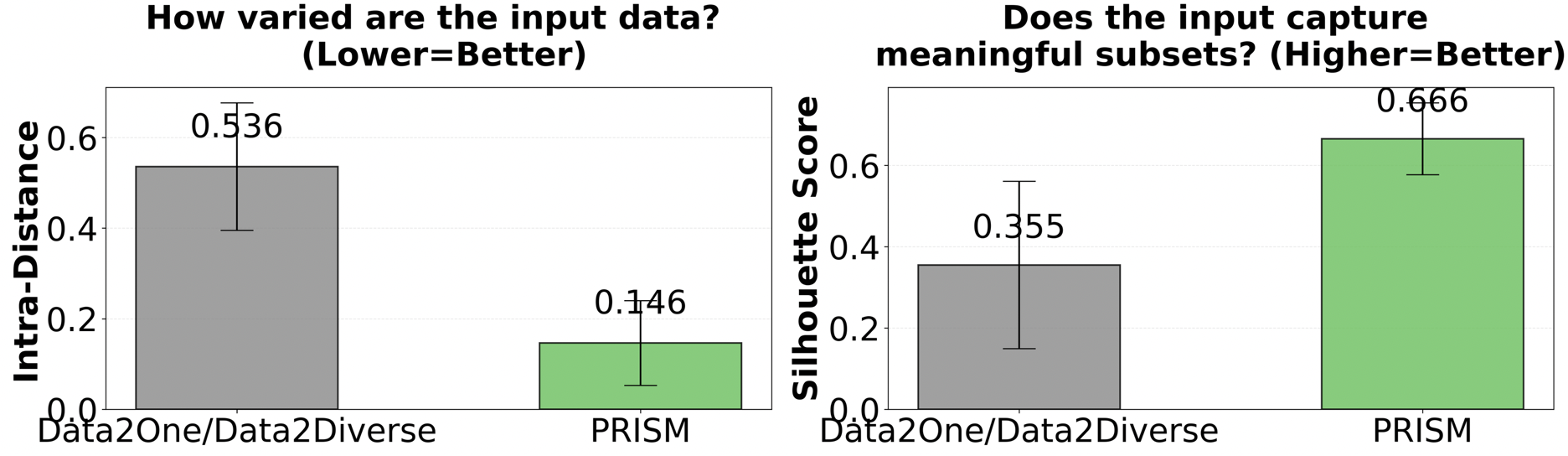}
    \caption{\textbf{Quantitative Analysis on Data-Drive Baselines' Low Fidelity.} (Left) computes the average GRAD distance between input examples. Lower intra-distance means that there is less variance in the input. (Right) We run K-medoids on each set of input before computing the silhouette score. A higher silhouette suggests that input contains more meaningful subsets.}
    \label{fig:fidelity_analysis_metric}
\end{figure}

\subsection{How important is the data curation step?\label{sec:exp_fidelity_discussion}}
We analyze why the data-driven baselines \dataOne{} and \dataDiv{} fail to outperform the baseline \cole{} in fidelity and why \ours{} avoids these pitfalls.
\dataOne{} and \dataDiv{} learn design knowledge from randomly sampled designs within a style.
Such sampling scheme creates two issues.
First, these samples have \emph{high variance}.
We quantify this variance using the average pairwise distance between example designs as a measure of this variance.
Fig.~\ref{fig:fidelity_analysis_metric} show that \ours{} has significantly lower average distances of $0.146$ compared to the baselines $0.536$. 
Note that \dataOne{} and \dataDiv{} have the same metric value because they have the exact same inputs, and Fig.~\ref{fig:fidelity_example} also provides an example of their inputs. 
Second, these samples \emph{do not guarantee meaningful clusters.}
We run K-medoids on each approach's input before calculating the silhouette score~\cite{rousseuw1987silhouettes}, which measures the cluster quality and higher is better.
Again, \ours{} achieve a higher silhouette score of $0.666$ compared to the baselines. 
These numerical differences manifest in the design knowledge that different approaches learn.
While \ours{} learns concrete, distinct patterns, \dataOne{} is only able to learn general guidelines due to the variance in the input set, while \dataDiv{} identifies superfluous subsets.
These findings highlight an important insight: data alone does not ensure improvement. Instead, careful incorporation of design data is essential for effective learning. 

\begin{figure}
    \centering
    \includegraphics[width=\linewidth]{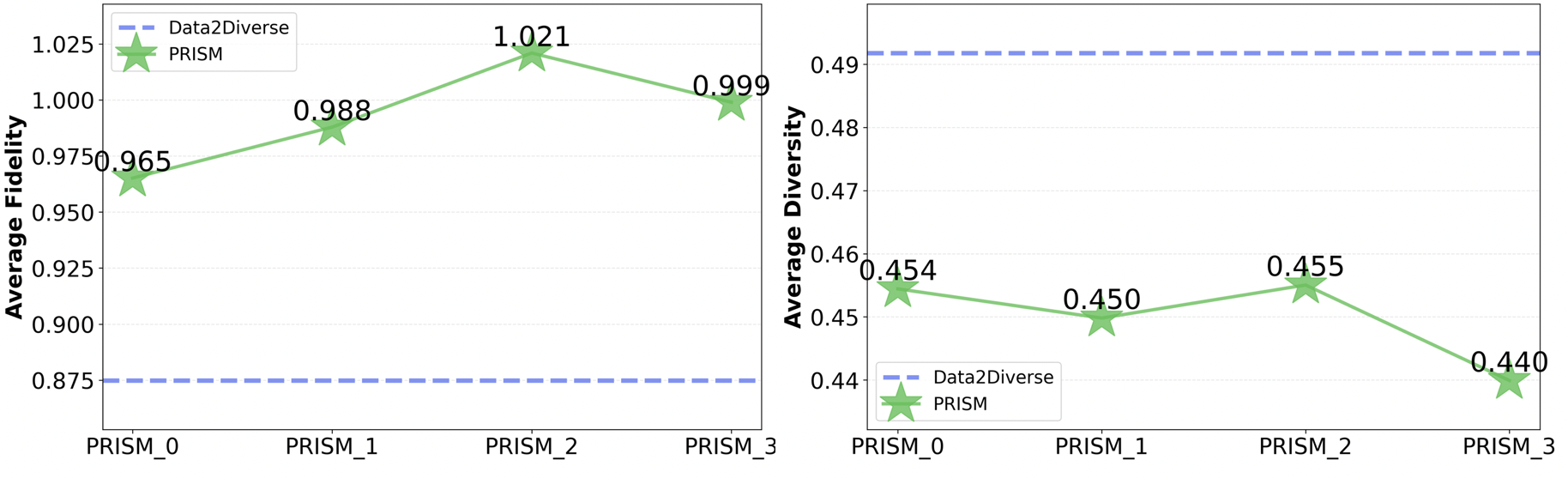}
    \caption{\textbf{Knowledge Refinement Results.} We ran 3 round of knowledge refinement. We report the average fidelity and diversity across 7 styles where \ours{}'s fidelity is close to baselines'.}
    \label{fig:improve}
\end{figure}

\subsection{How can the contrastive framework iteratively refine \ours{}' generated knowledge?}
During knowledge extraction, we can iteratively apply the contrastive framework to refine the knowledge learned by \ours{} and improves its alignment with the data.
We propose leveraging the positive and negative samples to evaluate the current knowledge's discriminative power and provide signals for refinement.
At a high level, the goal is to enhance the knowledge so that it accurately characterizes positive designs while remaining distinct from negative ones. 
For the set of $i$ positive and $j$ negative example design, we employ a VLM-based judge to evaluate whether each example is well described by the current knowledge. Feedback from false positives and false negatives is aggregated to guide subsequent refinement steps. Additional implemnetation details are in the Appendix. 

\noindent \textbf{Results. } We conduct experiments running 3 refinement iterations. As shown in Fig.~\ref{fig:improve}, fidelity generally increases with each iteration. Notably, all 3 iterations achieve higher fidelity compared to the original knowledge with an average fidelity of $0.965$. 
Meanwhile, the diversity values tend to fluctuate without any significant gain. 
We provide more in-depth discussion and a qualitative example illustrating the improvement process in the Appendix.

\begin{table}[t]
    \centering
    \small
    \setlength{\tabcolsep}{6pt}
    \renewcommand{\arraystretch}{1.1}
    \begin{tabular}{lcc}
        \toprule
        \textbf{Method} & \textbf{Fidelity} & \textbf{Diversity} \\
        \midrule
        PRISM                 & $0.999 \pm 0.032$ & $0.684 \pm 0.043$ \\
        \midrule
        PRISM (No Neg)   & $0.941 \pm 0.044$ & $0.684 \pm 0.047$ \\
        PRISM (No Individual) & $0.949 \pm 0.048$ & $0.657 \pm 0.038$ \\
        PRISM (No Collage)    &  $0.993 \pm 0.051$ & $0.642 \pm 0.044$ \\
        \midrule
        PRISM (No Prop-Retrieve) & $0.982 \pm 0.034$ & $0.658 \pm 0.043$ \\
        \bottomrule
    \end{tabular}
    \caption{\textbf{\ours{} Ablations}. We report the average fidelity and diversity across all 15 styles.}
    \label{tab:knowledge_ablation}
    \vspace{-1em}
\end{table}

\subsection{Ablation Studies}
\textbf{How to best utilize the design clusters during knowledge extraction?}
We ablate on the type of inputs  during the style knowledge extraction stage.
Because \ours{} learn design knowledge from a decent number of positive ($i=25$) and negative ($j=10$) designs, we empirically include the first 5 designs as individual examples while formatting the remaining examples as collages for positives and negatives respectively.
Thus, we explore
\ours{} (No Neg), which only provides positive examples,  
\ours{} (No Collage), which provides all examples as independent designs, and \ours{} (No Individual), which only provides collages as input. 
Table.~\ref{tab:knowledge_ablation} show that having the negative examples benefit fidelity as \ours{} and other ablations all have higher fidelity compared to the baseline \ours{} (No Neg) of $0.965$. 
Meanwhile, for diversity, having both the individual designs and collages seem to lead to slightly higher diversity compared to ones that remove one of the inputs. 

\noindent \textbf{What is the effect of proportional retrieval during inference?}
\ours{} sample design knowledge proportionally to the original design data distribution, so clusters with more designs are used more frequently. 
To evaluate this mechanism, we experiment on \ours (No Prop-Retrieve), which instead samples all appropriate design knowledge uniformly. 
Table.~\ref{tab:knowledge_ablation} show that proportional retrieval lead to both higher fidelity ($1.003 \rightarrow 1.018$) and diversity ($0.604 \rightarrow 0.630)$. 


\begin{figure}
    \centering
    \includegraphics[width=0.9\linewidth]{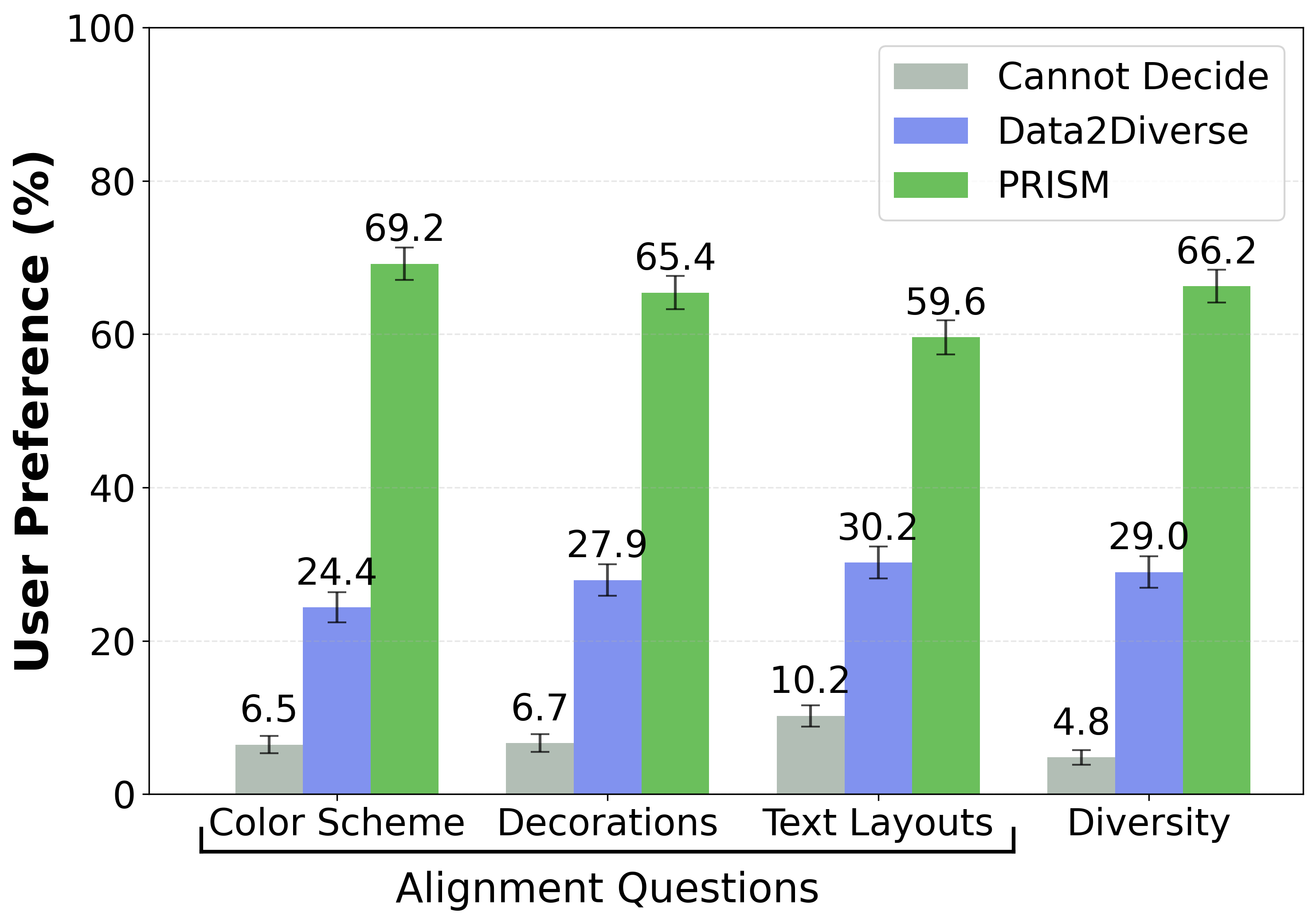}
    \caption{\textbf{Aggregated User Study Results.} We report the average user preference on each option. The first three questions ask the users which approach's design improvement aligns to the design style in terms of color scheme/decorations/text layouts. The last one ask the user which approach's output look more diverse.}
    \label{fig:user_main_results}
    \vspace{-1.5em}
\end{figure}

\section{User Study}
\noindent \textbf{Setup.}
To evaluate how well human preferences align with our quantitative metric, we conduct a user study involving 30 designers on the Prolific crowd-sourcing platform~\cite{prolific}. 
We investigate whether \ours{} can make design improvements that align with the design data compared to the most competitive baseline \dataDiv{} (as seen in Fig. \ref{fig:main_results}). 
Each designer is presented with a control and 15 test cases in random order, where each test case corresponds to one of the design styles.
For each test case, the designer is provided with the original design, a language instruction, two sets of design improvements from \ours{} and \dataDiv{} respectively (presented in random order, and blindly), and some reference examples to help the designer understand the design style. 
The designers are asked to compare the two approaches on whether the outputs align with the requested style across different design aspects: color scheme, decorations, text layouts. 
They also judge which approach's output looks visually more diverse. 
Details in the Appendix. 

\noindent \textbf{Results and Analysis.}
Fig.~\ref{fig:user_main_results} shows that designers consistently judged \ours{} as being more effective in creating designs that are more visually aligned to the requested design style. 
Designers also choose \ours{} $66.2\%$ of the times as the approach with more diverse improvements.
We further analyze styles where \dataDiv{}'s quantitative performance was close to \ours{}---such as ``colorful", ``dynamic", ``graphics"---shown in Fig.~\ref{fig:user_study_example} in the Appendix. Although quantitative metrics are similar, the user study reveals that \ours{} outperforms \dataDiv{} in stylistic alignment, except for color scheme in the ``colorful" style. This outcome is reasonable, as colorful designs are relatively easy to interpret without detailed data analysis. 
Similarly, users favor \ours{} for diversity in most cases except the ``floral" style, where outputs from both methods appear visually similar.


\section{Conclusion}
We investigate the problem of stylistic design improvement using VLMs, where the VLMs pretrained knowledge is often too general and lacks alignment with specific domain data.
Our key insight is to leverage existing design data to explicitly learn design knowledge.
We present \ours{} that first cluster the designs then employs a contrastive framework to learn distinctive design knowledge that captures different visual varieties. 
At inference time, \ours{} proportionally retrieves appropriate design knowledge so that the improvements reflect the original data distribution. 

We recognize a few limitations of \ours{}: (1) it currently treats the design data as a static set, whereas design data may evolve over time. Future work will explore efficiently strategies for incorporating new information and updating design knowledge; (2) the optimal representation of design knowledge may depend on the downstream module. We plan to investigate learning a design knowledge adapter that uses downstream execution signals to determine the most effective way to translate design knowledge for specific tasks.
{
    \small
    \bibliographystyle{ieeenat_fullname}
    \bibliography{main}
}

\clearpage
\setcounter{page}{1}
\maketitlesupplementary
\part{} 
\vspace{-1em}
\parttoc 

\section{Iterative Knowledge Refinement}
The knowledge extraction stage described in Sec.~\ref{sec:approach_knowledge_extraction} aims to extract compact, discriminative design knowledge for each cluster identified during the style space partitioning stage ( Sec.~\ref{sec:approach_data_curation}). 
By contrasting these positive and negative examples, the VLM synthesizes design knowledge that captures a unique type of visual characteristics within a style. 
This contrastive framework naturally extends to provide feedback signals for iteratively improving the extracted knowledge.

\subsection{Overall Approach}
The goal of knowledge refinement is to refine the learned knowledge so that it is more data-aligned and discriminative.
Thus, we introduce iterative knowledge refinement, which leverages the contrastive signals to identify weaknesses and suggest improvements. 
Specifically, the process involves detecting misaligned designs, generating feedback, and refining the knowledge accordingly. 

Let $k_0$ denote the initial knowledge for a cluster. Let $\mathcal{D^+} = \{X^+\}$ be  the set of positive designs belonging to this cluster and $\mathcal{D^-} = \{X^-\}$ be the set of negative designs from other clusters. Algorithm~\ref{alg:refine} shows the overall computation.

\noindent \textbf{Identify Misaligned Designs.} To identify where the current knowledge fails to align to the data, we introduce a \emph{Design Classifier} $c(k, X)$ that determines whether a design $X$ aligns with the knowledge $k$. 
Ideally, a discriminative knowledge $k$ should align with designs that are positive examples (so $\forall X^+ \in D^+, c(k, X^+)=1$) and distinguish designs that are negative examples (so $\forall X^- \in D^+, c(k, X^-)=0$). We run classifiers on all examples to identify the false negatives and false positives. Implementation details are provided in Sec.~\ref{sec:app_refine_imp_details}. 

\noindent \textbf{Generate Feedback.} Given the set of false negatives and false positives, we collect feedback on how to improve the current knowledge. Thus, we define a \emph{Feedback Generator} $f(k, X)$ that analyzes why the design is misclassified before generating actionable advices on how to improve the design. We accumulate all the feedback as $\mathcal{H}$ for the next step. 

\noindent \textbf{Refine Knowledge.} Given all the feedback $\mathcal{H}$, the goal of the \emph{Knowledge Refiner} $r(k, H)$ is to analyze all the feedback before editing the current knowledge. This newly refined knowledge is used in the next iteration. 

\begin{algorithm}[tb]
\caption{\ours{} Iterative Knowledge Refinement}
\label{alg:refine}
\begin{algorithmic}
    \STATE {\bfseries Input:} Initial Knowledge $k_0$, Positive Examples $\mathcal{D^+}$, Negative Examples $\mathcal{D^-}$, Refinement Iterations $T$, Design Classifier $c(k, X)$, Feedback Generator $f(k, X)$, Knowledge Refiner $r(k, \mathcal{H})$
    \FOR{$t=0$ {\bfseries to} $T$}
        \STATE Feedback History $\mathcal{H}_t \gets \emptyset$
        \STATE \algcommentlight{Collect feedback from false negatives}
        \STATE $\text{FN}_t = \{ X^+ \in \mathcal{D}^+ \mid c(k_t, X^+) = 0 \}$
        \STATE $\mathcal{H}_t \gets \mathcal{H}_t \cup \{ f(k_t, X^+) \mid X^+ \in \text{FN}_t \}$
        \STATE \algcommentlight{Collect feedback from false positives}
        \STATE $\text{FP}_t \gets \{ X^- \in \mathcal{D}^- \mid c(k_t, X^-) = 1 \}$
        \STATE $\mathcal{H}_t \gets \mathcal{H}_t \cup \{ f(k_t, X^-) \mid X^- \in \text{FP}_t \}$
        \STATE \algcommentlight{Refine knowledge using feedback}
        \STATE $k_{t+1} \gets r(k_t, \mathcal{H}_t)$
    \ENDFOR
    \STATE {\bfseries Return $k_T$} 
\end{algorithmic}
\end{algorithm}

\begin{figure*}[ht]
    \centering
    \includegraphics[width=\textwidth]{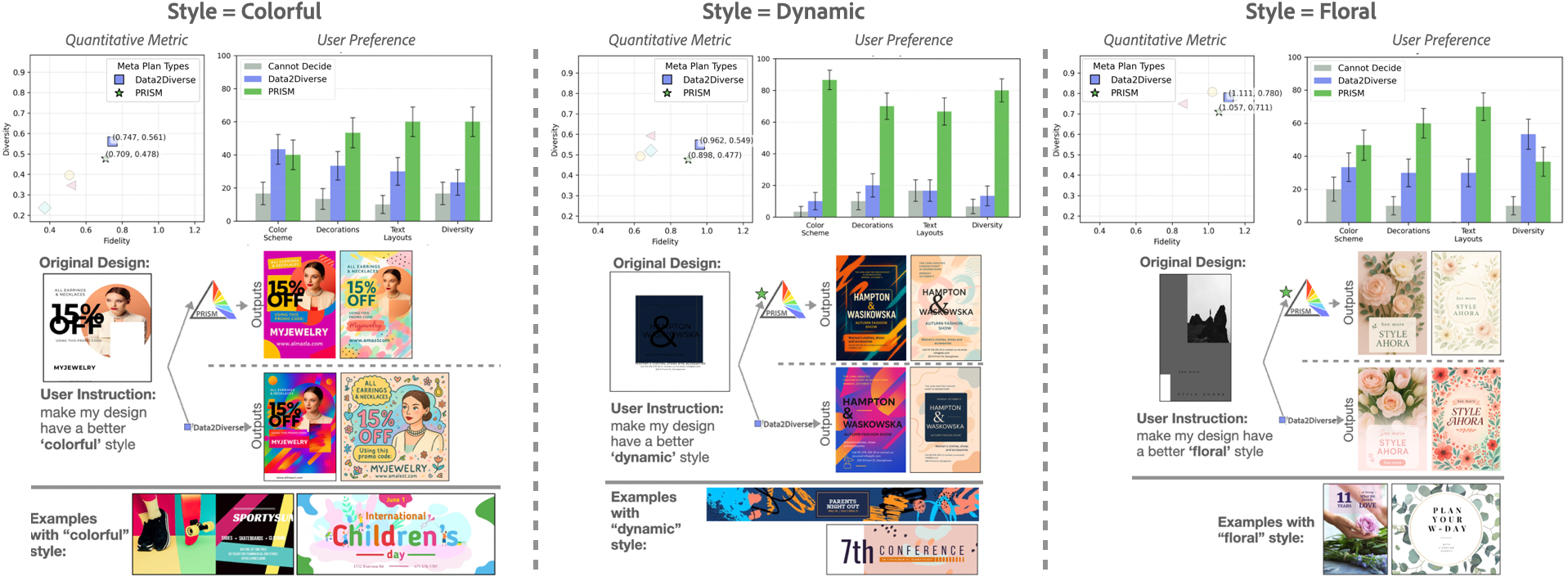}
    \caption{\textbf{User Study Qualitative Examples.} For each style, we show the quantitative metric as well as the average user preferences across different questions. Below each shows the original design, the language instruction, two approach's output (during the study, the user does not know which output corresponds to what approach), and representative examples of a style.}
    \label{fig:user_study_example}
\end{figure*}

\subsection{Implementation Details\label{sec:app_refine_imp_details}}
\textbf{Design Classifier. } To access how well a learned design knowledge aligns with a design, we formulate the task as a multi-class classification problem by asking: which design knowledge does a design align with the most strongly? Specifically, each cluster represents a class and has an associated learned design knowledge.
Naively, we can frame it as a multiple choice problem where we ask a VLM to choose among a set of design knowledge which one is best describes a design. 
However, as prior works show that pairwise comparison is more reliable~\cite{jeong2025comparativetrappairwisecomparisons}, we also leverage pairwise comparison to perform multi-class classification.
Specifically, for each pair of design knowledge, we prompt a VLM to decide which one best describes a design.
The design knowledge with the highest amount of pairwise wins is the classification label (tie-breaking randomly). 
Then, the classifier $c(k,X) = 1$ if a design is classified to align the most with the knowledge $k$ and $c(k,X) = 0$ if a design aligns the most with knowledge other than $k$ that correspond to other clusters.

\noindent \textbf{Hyperparameter. } We use \texttt{gpt-4.1} as the VLM, querying it with temperature $0.3$.

\begin{figure*}[ht]
    \centering
    \includegraphics[width=\textwidth]{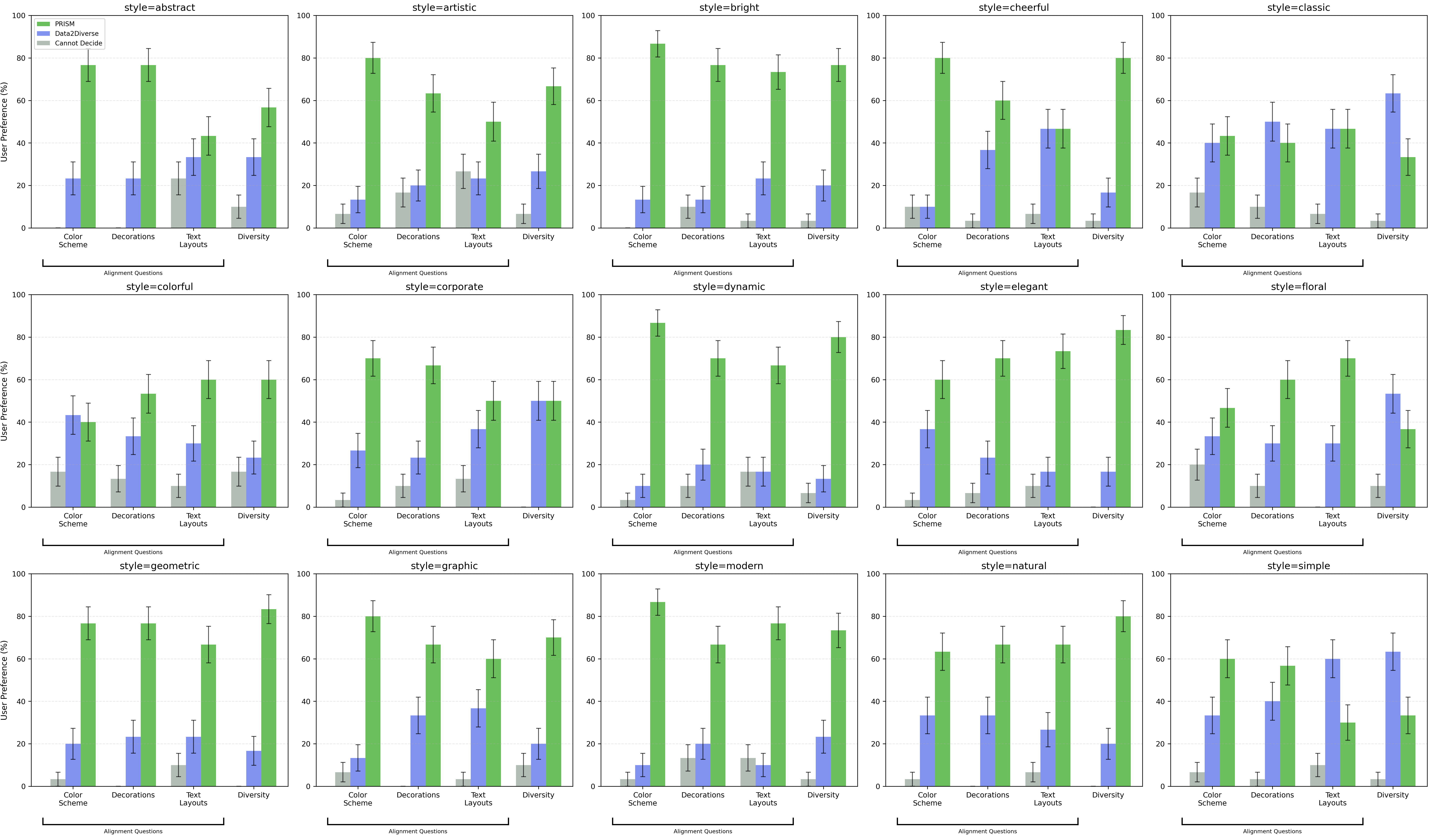}
    \caption{\textbf{User Study Per-Style Preferences.} We report the average user preferences for each design style.}
    \label{fig:user_per_style}
\end{figure*}
\section{User Study}
\subsection{Setup}
The goal is to evaluate how well do designer preferences align with our quantitative metric by specifically comparing \ours{} against the most competitive baseline \dataDiv{}.
For each test case (each corresponds to a design style), the user is provided with the original design to improve, language instruction, two approaches' output, and reference examples to help the design understand the design style.
Example test cases are shown in Fig.~\ref{fig:user_study_example}.
For each style, we select the specific outputs where the two approaches are close in their raw fidelity score. 
Each test case contains the following questions with options ``A is Better", ``B is Better", ``Cannot Decide":
\begin{enumerate}
    \item Which approach's outputs are more aligned with \textbf{the color scheme} of the example \{design\_style\} designs?
    \item Which approach's outputs are more aligned with \textbf{the decorations} of the example \{design\_style\} designs?
    \item Which approach's outputs are more aligned with the \textbf{text effects} of the example \{design\_style\} designs?
    \item Which approach's outputs look more diverse?
\end{enumerate}

\subsection{Per-Style Results}
We report the per-style user preferences in Fig.~\ref{fig:user_per_style}. 
Among all the styles, $11/15$ have the users preferring \ours{} over \dataDiv{} across all questions. $3/15$ (``colorful", ``floral", ``simple") have the users preferring the baseline \dataDiv{} on one or two questions. $1/15$ (``classic") have the users preferring the baseline \dataDiv{} on all questions.  

\begin{figure*}[ht]
    \centering
    \includegraphics[width=\textwidth]{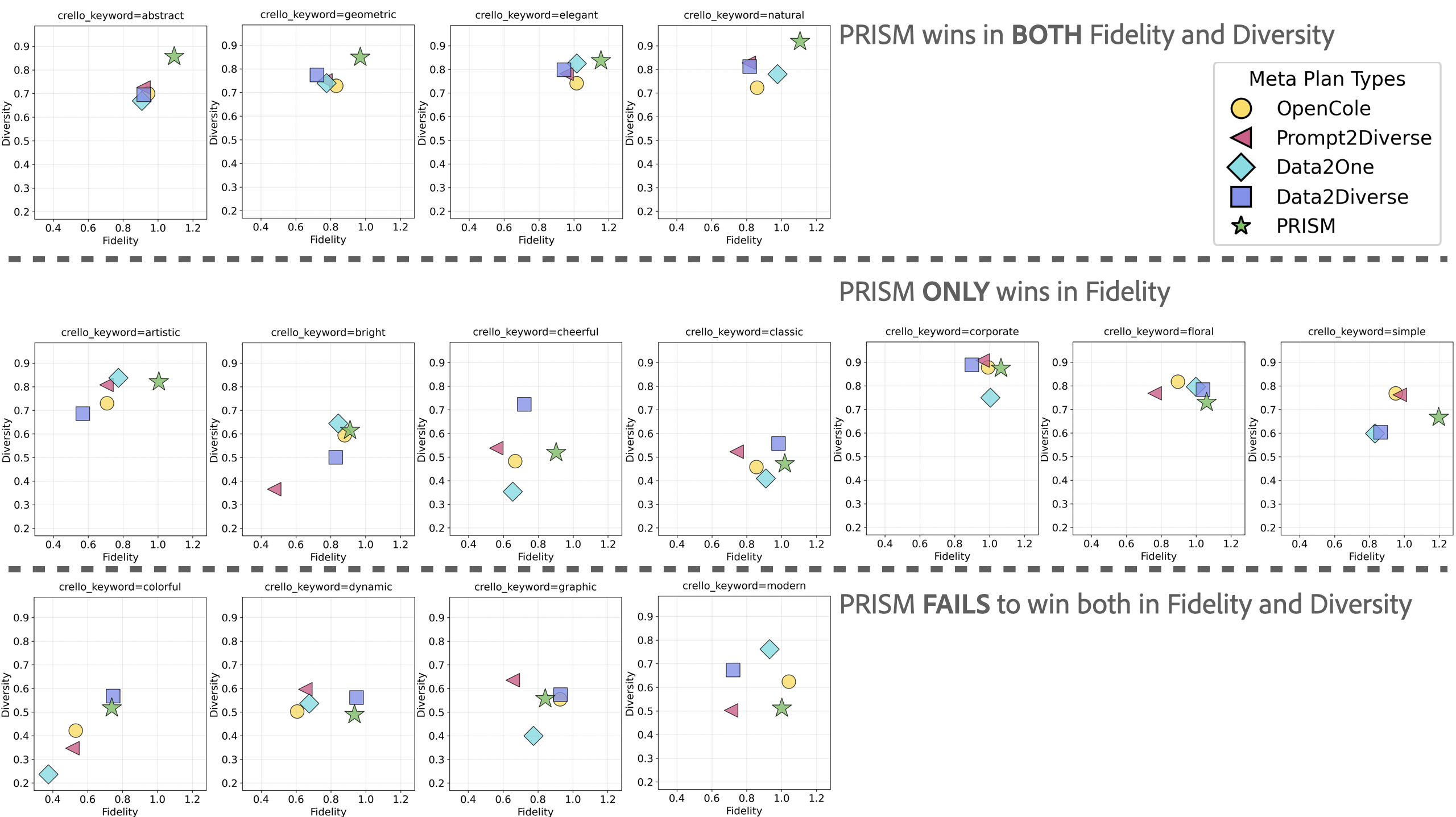}
    \caption{\textbf{Per-Style Average Fidelity and Diversity.} We report the bootstrapped average fidelity and diversity for each of the 15 styles. We categorize the styles into 3 categories: (1) \ours{} has the highest fidelity and diversity (2)\ours{} only has the highest fidelity (but not diversity) (3) other baselines have the highest fidelity and diversity.}
    \label{fig:per_style_results}
\end{figure*}
\begin{figure*}[ht]
    \centering
    \includegraphics[width=\textwidth]{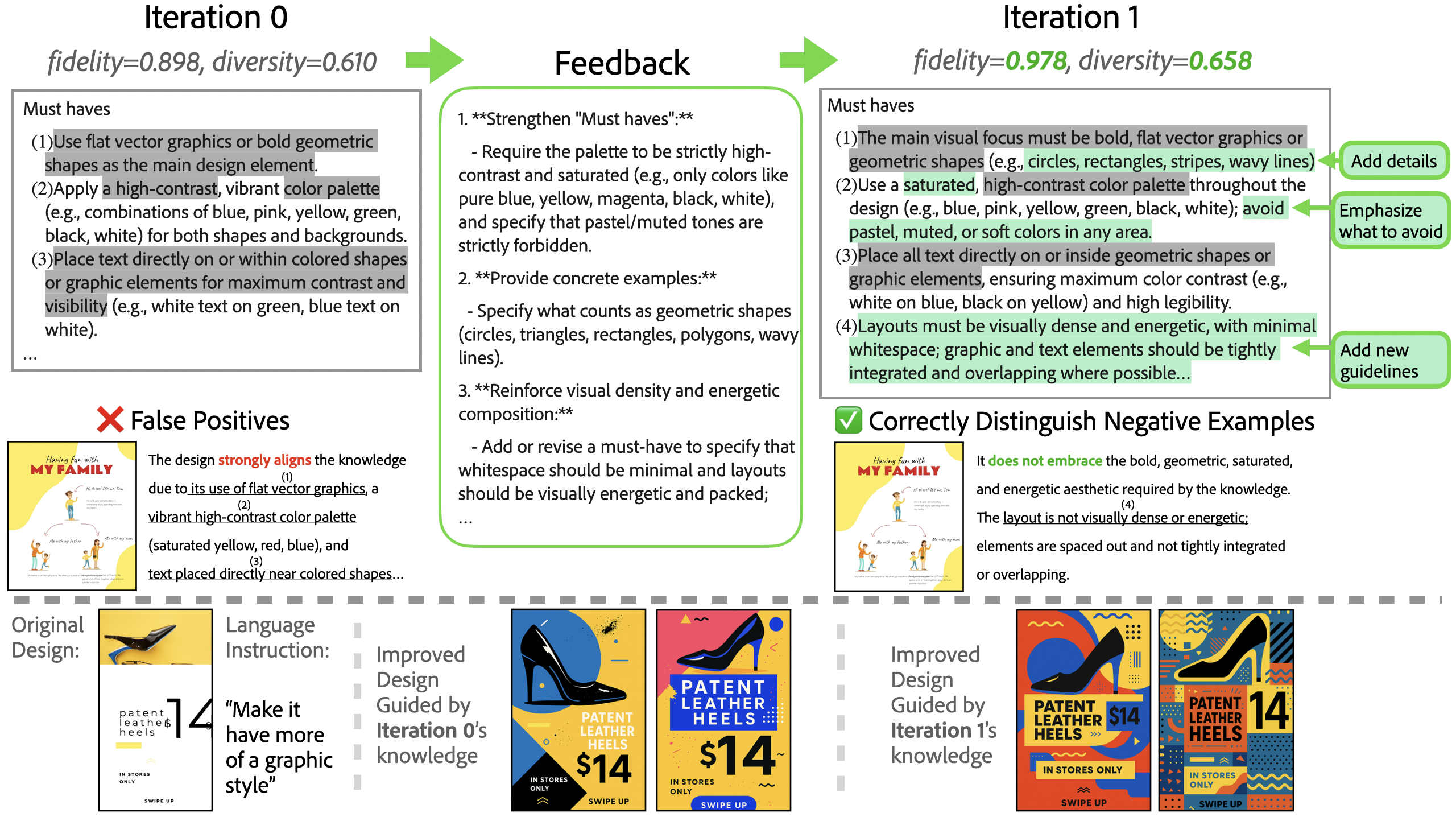}
    \caption{\textbf{Iterative Knowledge Refinement Qualitative Example.} Left shows design knowledge learned during \ours{} iteration 0. It fails to distinguish a negative example. From these failure cases, we generate feedback that is used to refine the knowledge. Right shows iteration 1's refined knowledge, where additional details and new guidelines are added. Iteration 1 succeeds in distinguish the same negative example. Bottom row shows qualitative example of improved designs guided by knowledge from different iterations. For the two design knowledge, similarities are highlighted in gray, while edits are highlighted in green.}
    \label{fig:iter_refine_example}
\end{figure*}

\section{Quantitative Experiments}
\subsection{Design Data}
We combine the train set (19479) and the validation set (1852) from the Crello~\cite{yamaguchi2021canvasvae} dataset to create the overall pool of design. 
The first step is to identify the design style for each design. 
Each design has a list of associated keywords, where some are about the topics and others are about styles (e.g., ``quote, bright, colorful, abstract, graphic, font, t-shirt, tshirt, typography"). 
We use NLTK's~\cite{bird-loper-2004-nltk} part of speech tagger to identify keywords that are adjectives before using an LLM (\texttt{gpt-4.1}) to identify adjectives that describe a design style.
The second step is to identify design styles that have sufficient amount of data: we only keep design styles that have more than 100 designs.
The third step is to filter out repetitive designs within each style.
Given all the designs that share the design style, we keep the largest designs among ones that have the same title (because they are just different cropped version of the same content) and we also keep the largest designs among ones that have close average perceptual hash (differences less than 10).
In the end, we acquired 15 design styles: abstract, artistic, bright, cheerful, classic, colorful, corporate, dynamic, elegant, floral, geometric, graphic, modern, natural, simple. 

\subsection{Implementation Details}
Design knowledge is generated at temperature $0.3$.
During inference, given a design to improve and the language instruction, the general pipeline is to generate a design plan before generating design improvements conditioning on that plan. For baselines generates zero or only one design knowledge (\cole{}, \dataOne{}), we set the design plan generation temperature to $0.7$ to encourage more diversity. Meanwhile, approaches with more than one design knowledge (\promptDiv{}, \dataDiv{}, \ours{}) generates design plan at a lower temperature of $0.3$. Image generation does not have a temperature hyperparameter.

\subsection{Per-Style Results}
We present per-style results in Fig.~\ref{fig:per_style_results}. PRISM has the highest fidelity on $11/15$ styles and the highest diversity on $5/15$ styles, matching the results when we compute the expected ranks across all styles in Fig.~\ref{fig:rank}.

\subsection{Iterative Knowledge Refinement Details}
We study the effect of knowledge refinement on styles where \ours{} fails to have the highest fidelity and diversity (colorful, dynamic, graphic, modern) and where \ours{}'s value is close to the baselines (bright, classic, floral). 
Fig.~\ref{fig:iter_refine_example} shows the qualitative example. The design knowledge from the \ours{}'s Iteration 0 lacks sufficient details, making it misclassify a negative example as belonging to the cluster that it represents.
Using information about false positive, we generate feedback that aims to make the design knowledge more concrete and discriminative.
The resulting design knowledge in Iteration 1 achieves higher fidelity ($0.898 \rightarrow 0.978$) and slightly higher diversity ($0.610 \rightarrow 0.658$). 
With the new details, emphasis, and new guidelines, this new refined knowledge successfully distinguishes the negative example, where Iteration 0 fails.

\end{document}